\documentclass[runningheads,a4paper,nonatbib]{llncs}[2015/06/24]

\usepackage{cmap}
\usepackage[T1]{fontenc}

\usepackage{graphicx,subfig}
\usepackage{amsmath}
\usepackage{amsfonts}       
\usepackage[ngerman,english]{babel}
\addto\extrasenglish{\languageshorthands{ngerman}\useshorthands{"}}

\usepackage[%
rm={oldstyle=false,proportional=true},%
sf={oldstyle=false,proportional=true},%
tt={oldstyle=false,proportional=true,variable=true},%
qt=false%
]{cfr-lm}
\usepackage[math]{blindtext}

\usepackage{cite}

\usepackage{paralist}

\usepackage{csquotes}

\usepackage{microtype}

\usepackage{url}
\makeatletter
\g@addto@macro{\UrlBreaks}{\UrlOrds}
\makeatother

\usepackage{xcolor}

\usepackage{pdfcomment}
\hypersetup{hidelinks,
   colorlinks=true,
   allcolors=black,
   pdfstartview=Fit,
   breaklinks=true}
\usepackage[all]{hypcap}

\usepackage[capitalise,nameinlink]{cleveref}
\crefname{section}{Sect.}{Sect.}
\Crefname{section}{Section}{Sections}

\usepackage{xspace}

\DeclareFontFamily{U}{MnSymbolC}{}
\DeclareSymbolFont{MnSyC}{U}{MnSymbolC}{m}{n}
\DeclareFontShape{U}{MnSymbolC}{m}{n}{
    <-6>  MnSymbolC5
   <6-7>  MnSymbolC6
   <7-8>  MnSymbolC7
   <8-9>  MnSymbolC8
   <9-10> MnSymbolC9
  <10-12> MnSymbolC10
  <12->   MnSymbolC12%
}{}
\DeclareMathSymbol{\powerset}{\mathord}{MnSyC}{180}

\begin{document}

\title{Multi-Resolution Networks for Semantic Segmentation in Whole Slide Images}
\titlerunning{MRN for Semantic Segmentation in WSIs}

\author{Feng Gu \and Nikolay Burlutskiy \and Mats Andersson \and Lena Kajland Wil\'en}
\authorrunning{Feng Gu et al.}
\institute{ContextVision AB, Link\"oping, Sweden\\\email{feng.gu@contextvision.se}}

%

\maketitle

\begin{abstract}
Digital pathology provides an excellent opportunity for applying fully convolutional networks (FCNs) to tasks, such as semantic segmentation of whole slide images (WSIs). However, standard FCNs face challenges with respect to multi-resolution, inherited from the pyramid arrangement of WSIs. As a result, networks specifically designed to learn and aggregate information at different levels are desired. In this paper, we propose two novel multi-resolution networks based on the popular `U-Net' architecture, which are evaluated on a benchmark dataset for binary semantic segmentation in WSIs. The proposed methods outperform the U-Net, demonstrating superior learning and generalization capabilities.
\end{abstract}


\section{Introduction}\label{sec:intro}

The working pattern of an experienced pathologist is frequently characterized by a repeated zooming in and zooming out motion while moving over the tissue to be graded. This behavior is similar if a microscope is used or if a whole slide image (WSI) is observed on a screen. The human visual system needs these multiple perspectives to be able to grade the slide. Only in rare cases can a local neighborhood of a slide be safely graded, independent from the surroundings. The heart of the matter is that the slide only represents a 2D cut out of a complex 3D structure. A glandular tissue in 3D resembles the structure of a cauliflower. Depending on the position of the 2D cut, the size and shape of the glands on the slide may vary significantly. To assess if a deviation in size or shape is due to the position of the cut or to a lesion, a multi-resolution view is crucial. The structure of the surrounding glands must be accounted for when the current gland is being investigated at a higher resolution.

Digital pathology opens the possibility to support pathologists by using fully convolutional networks (FCNs)~\cite{long2015} for semantic segmentation of WSIs. Standard FCNs do however face the same challenge with respect to multi-resolution. It can be argued that a network like U-Net~\cite{ronneberger2015} can to some extent handle multi-resolution, since such a structure is inherent in the network. However, patches (image regions of a WSI at a given resolution) with the finest details should probably be extracted at the highest resolution, to utilize such a capability. It may also require the patch size to be considerably larger, making it infeasible for the VRAM of a modern GPU to fully explore the multi-resolution. As a result, approaches capable of learning from data and aggregating information efficiently and effectively at multiple resolutions are desired.

In this paper, two novel multi-resolution networks are proposed to learn from input patches extracted at multiple levels. These patches share the same centroid and shape (size in pixels), but with an octave based increase of the pixel size, micrometers per pixel (mpp). Only the central high resolution patch is segmented at the output. The proposed methods are evaluated and compared with the standard U-Net on a benchmark dataset of WSIs.

\section{Related Work}\label{sec:rl_work}

Semantic segmentation problems were initially solved by traditional machine learning approaches, where hand crafted features were engineered \cite{segmentation_ml16}. Researchers applied methods, such as predictive sparse decomposition and spatial pyramid matching, to extract features of histopathological tissues \cite{spm_13}. However, deep learning approaches based on FCNs \cite{long2015} showed significantly higher performance and eventually have substituted them \cite{review_deep_seg17}. To overcome the so called `checkerboard artifacts' of transposed convolutions, several approaches have been proposed, e.g. SegNet \cite{segnet15}, DeepLab-CRF \cite{DCNN_CRF14}, and upscaling using dilated convolutions \cite{yu2017}. To increase localization of learned features, high resolution features from the downsampling path can be aggregated with the upsampled output. Such an operation is known as `skip connections', which enables a successive convolution layer to learn and assemble a more precise output based on the aggregated information. Several researchers successfully demonstrated that architectures with skip connections can result in better performance. Such networks include U-Net, densely connected convolutional networks \cite{densenet16}, and highway networks with skip connections \cite{highway_networks15}. On overall, U-Net has proved to be one of the most popular networks for biomedical segmentation tasks \cite{lung_segmentation18}.

One limitation of standard FCNs is the fact that the networks are composed of convolution layers with a set of filters that have the same receptive field size. The receptive field size corresponds to the context that a network can learn from, and eventually influences the network performance. Grais {\it et al.}~\cite{emad_17} proposed a multi-resolution FCN with different receptive field sizes for each layer, for the audio source separation problem. Such a design allowed to extract features of the same input at multiple perspectives (determined by the receptive field sizes), and thus to capture global and local details from the input. Fu {\it et al.} introduced a multi-scale M-Net \cite{fu2018} to tackle the problem of joint optic disc and cup segmentation, where the same image contents of different input shapes or scales are passed through the network. However, both methods were designed to handle the same input audio or image content, while learning features from multiple perspectives by either employing encoders with varied respective fields or taking inputs with multiple scales. Roullier {\it et al.}~\cite{multires_ROULLIER2011} proposed multi-resolution graph-based analysis of whole slide images for mitotic cell segmentation. The approach is based on domain specific knowledge, which cannot be easily transferred to another problem domain. Recently, an approach of using multi-resolution information in FCN was described in \cite{miltires_Sachin_17}. However, the fusion of multi-resolution inputs was performed before encoders, instead of within the network. In addition, the approach could only be applied to a subset of small regions of interests, rather than WSIs. 

Networks that incorporate inputs extracted from different resolutions with respect to the same corresponding tissue area in WSIs are desired, to tackle the challenge of multi-resolution effectively. In addition, the networks should be scalable in terms of resolutions and VRAM efficient for training and prediction. These motivated us to develop the multi-resolution networks in this work.

\section{Algorithmic Formulation}\label{sec:algo_form}

A common practice of handling a WSI with deep learning is to divide it into multiple equally sized patches~\cite{wang2016}. Here the deep learning task is formulated as a binary semantic segmentation problem of patches, where each patch is considered an image example. At prediction, a trained model first predicts each patch individually, and then stitches predictions of all the patches, to form the prediction of the entire slide (a probabilistic map indicating the probability of each pixel belonging to the class of interest).

\subsection{Learning and Inference}

Let $(\mathbf{x}, y)\in\mathbf{X}\times\mathbf{Y}$ be a patch or an example of a given dataset, where $\mathbf{X}\subseteq\mathbb{R}^{N\times D\times 3}$ and $\mathbf{Y}\subseteq\mathbb{N}^{N\times D}$. The value of $N$ is equal to the number of examples (or patches), and $D$ is the dimensionality of the feature vector (i.e. the product of height and width of the patch $h\times w$). So $\mathbf{x}$ can be a RGB image extracted from a slide, and $y$ can be a binary ground truth mask associated with the RGB image. We can formulate a deep network as a function $f(\mathbf{x};\mathbf{W})$, where $\mathbf{W}$ is a collection of weights of all the parametrized layers. The learning task is a process of searching for the optimal set of parameters $\hat{\mathbf{W}}$ that minimizes a loss function $\mathcal{L}(y, f(\mathbf{x};\mathbf{W}))$. The output of the function $f$ can be transformed to a probabilistic value in the range of $[0,1]$ via a sigmoid function. A commonly used loss function for binary semantic segmentation is the binary cross entropy loss.

To counter over-fitting and improve the generalization capability of a trained model, a regularization term $\mathcal{R}(\cdot)$ is often added to the objective function as
\begin{equation}
\label{eqn:objective_fun}
\mathcal{E}_{\mathbf{W}}=\sum_{i=1}^{N}\mathcal{L}\left(y, f(\mathbf{x}_{i};\mathbf{W})\right)+\lambda\mathcal{R}(\mathbf{W})
\end{equation}
where the scalar $\lambda$ determines the weighting between two terms. One popular regularization function is $\ell_{2}$-regularization, such that $\mathcal{R}(\mathbf{W})=\|\mathbf{W}\|_{2}^{2}$. Search of the optimal set of parameters $\hat{\mathbf{W}}=\arg\min_{\mathbf{W}}\mathcal{E}_{\mathbf{W}}$ for the objective function is known as optimization in machine learning. Popular optimizers include stochastic gradient descent (SGD), adaptive gradient (AdaGrad), and root mean square propagation (RMSProp). Recently, adaptive moment estimation (Adam)~\cite{kingma2015} has become a particularly popular method for optimizing deep networks. 

\subsection{Multi-Resolution Networks}
Here we propose two multi-resolution networks (MRN) that are based on the architecture of U-Net~\cite{ronneberger2015}. A standard U-Net can be seen as two parts, an `encoder' for downsampling and a `decoder' for upsampling. The downsampled feature maps are concatenated with the corresponding layers of the decoder in the upsampling pathway. The proposed MRN employ multiple encoders corresponding to different resolutions that are structurally identical for downsampling, and one single decoder for upsampling.
\begin{figure}[t]
\centering
\includegraphics[width=\textwidth]{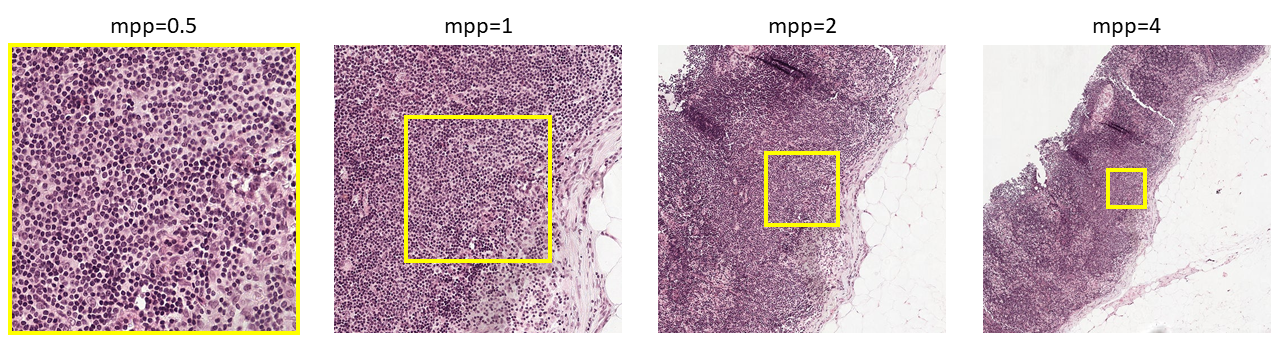}
\caption{From left to right are the patches with the same central coordinates, where mpp=0.5 is equivalent to 20x and so forth. The increase of mpp values corresponds to the zooming out action to enlarge the field of view, and the yellow squares represent the effective tissue ares at different magnifications.}
\label{fig:input_plots}
\end{figure}

\begin{figure}[t]
\centering
\includegraphics[width=0.33\textwidth, angle=90]{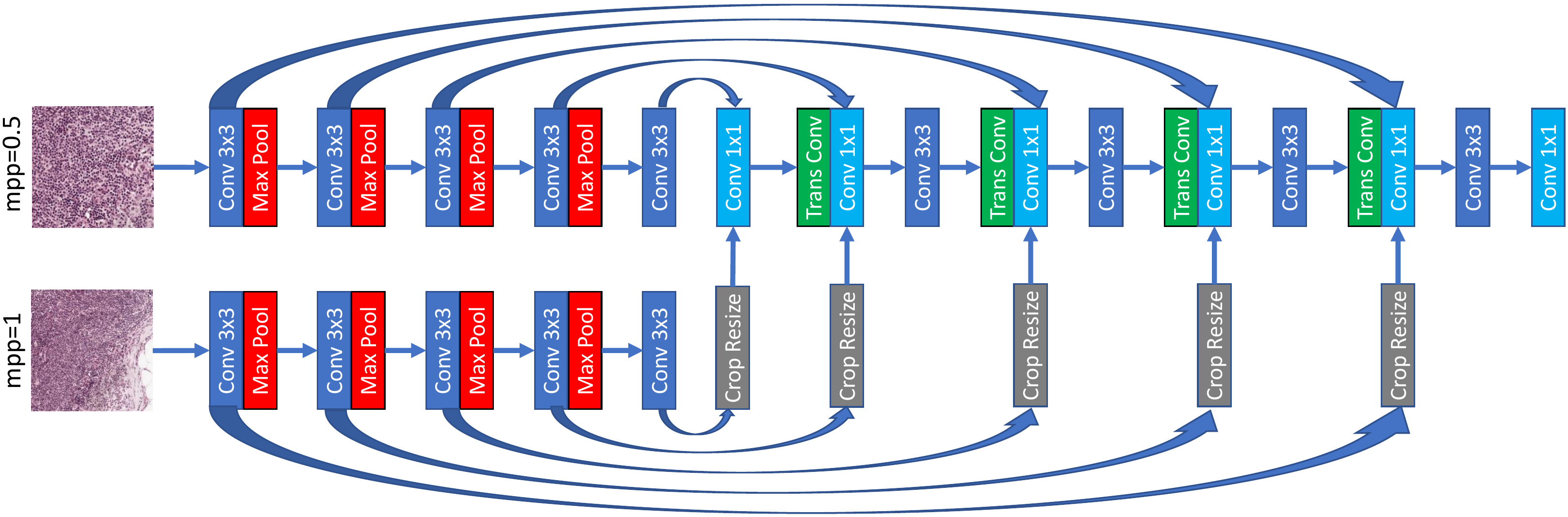}
\caption{An illustration of the proposed MRN methods when two resolutions are involved. The dark blue boxes represent stacks of two $3\times 3$ convolution layers with ReLU activations; the red boxes are $2\times 2$ max pooling layers; the light blue boxes are $1\times 1$ convolution layers with identity activations; the green boxes are $2\times 2$ transposed convolution layers with stride=2 and ReLU activations.}
\label{fig:mrn_unet}
\end{figure}

The input shapes of all resolutions are identical, and the examples share the common central coordinates and effectively cover tissue areas in a pyramid manner, as in Fig.~\ref{fig:input_plots}. Let $(\mathbf{x}, y)$ be an example, where $\mathbf{x}_j=[\mathbf{x}_1,\mathbf{x}_2,\ldots,\mathbf{x}_J]$ and $y=y_{1}$, where the resolutions are in a descending order. The shapes of $\mathbf{x}$ and $y$ are there $h\times w\times 3 \times J$ and $h\times w\times 1$ respectively. The rationale behind such an arrangement is that the pixel correspondence is more cumbersome compared to a standard U-Net. A key issue is to enable a sufficient receptive field for the low resolution branches of the network to successfully convey the information from the peripheral regions into the central parts.

To preserve the information relevant to the area of interest (i.e. the central part) at a lower resolution, we center crop the output feature maps of each encoder unit and then resize them back to the original resolutions via upscaling. We can defined a nested function $u\circ v$ such that
\begin{equation}
u: \mathbb{R}^{w\times h\times c}\rightarrow \mathbb{R}^{\lfloor\frac{w}{\gamma}\rfloor\times \lfloor\frac{h}{\gamma}\rfloor\times c}\quad\mathrm{and}\quad v: \mathbb{R}^{\lfloor\frac{w}{\gamma}\rfloor\times \lfloor\frac{h}{\gamma}\rfloor\times c}\rightarrow\mathbb{R}^{w\times h\times c} 
\label{eqn:center_crop_fun}
\end{equation}
where the cropping factor is $\gamma=2^{\mathbb{N}}$, since resolutions at different levels of a WSI are usually downsampled by a factor 2 in both height and width. On one hand, the function $u$ center crops a real-valued tensor of shape $h\times w\times c$ (height, width, and channels) to the shape of ${\lfloor\frac{w}{\gamma}\rfloor\times \lfloor\frac{h}{\gamma}\rfloor\times c}$. On the other hand, the function $v$ upscales the output of $u$ to the original shape. For upscaling, we present two options, namely `MRN-bilinear' via bilinear interpolation and `MRN-transposed' through transposed convolution.

The outputs of $u\circ v$ are concatenated with the convoluted feature maps of the corresponding layers in the encoder of the highest resolution. The concatenated feature maps are then passed though a $1\times 1$ convolution layer with an identity activation, before being combined with layers in the decoder. The $1\times 1$ convolution acts as a weighted sum to aggregate feature maps from all the resolutions, while keeping the number feature maps in the decoder constant despite the number of resolutions involved. Fig.~\ref{fig:mrn_unet} illustrates an example of such networks when two resolutions are involved, and this is easily expandable with more resolutions.

\section{Experimental Conditions}\label{sec:exp_conditions}

\subsection{Implementation Details}

We implemented all the networks in TensorFlow, with `SAME' padding. Batch normalization and $\ell_{2}$-regularization with $\lambda=0.005$ were applied to all the convolution and transposed convolution layers, to improve convergence rates and counter over-fitting. We employ the Adam optimizer with default parameters ($\eta$=0.001, $\beta_{1}$=0.9, $\beta_{2}$=0.999, and $\epsilon=10^{-8}$). The input shape is 512 in height and width, and the collection of resolutions are $\mathrm{mpp}\in\{0.5, 1, 2, 4\}$, where the U-Net deals with one of the resolutions at each time and the MRN methods handles all the resolutions simultaneously. The batch size is equal to 16, which is limited by the VRAM of an NVIDIA Titan XP. The number maximum epochs is set to 500 for the training to be terminated.

\subsection{Segmentation Experiments}
CAMELYON datasets~\cite{bejnordi2017} are the only few publicly available WSI datasets with pixel-level annotations. In particular, the CAMELYON16 dataset has both the training and testing sets available, and is one of the most popular benchmark datasets in the field of digital pathology. As a result, it was chosen for evaluating the proposed methods against the standard U-Net, for binary semantic segmentation of `normal' and `tumor' classes in WSIs \footnote{Note we have no intention to tackle the CAMELYON16 tasks of slide-based or lesion-based classifications, or the CAMELYON17 task of determining the pN-stage for a patient. Those tasks are beyond the scope of this work.}. There are 269 slides in the training set, 159 of which are normal and the remaining 110 are tumor. As pointed out in~\cite{wang2016}, 18 tumor slides have non-exhaustive annotations and thus are excluded from the experiments. The training set is then randomly divided into `training' (80\%) and `validation' (20\%), where the validation set is used to select the best model with respect to lowest validation losses. The testing set has 130 slides, 80 of which are normal and the rest are tumor. We excluded 2 tumor slides, due to non-exhaustive annotations. 
\begin{figure}[t]
\centering
	\subfloat[Validation Set]{\includegraphics[width=0.5\textwidth]{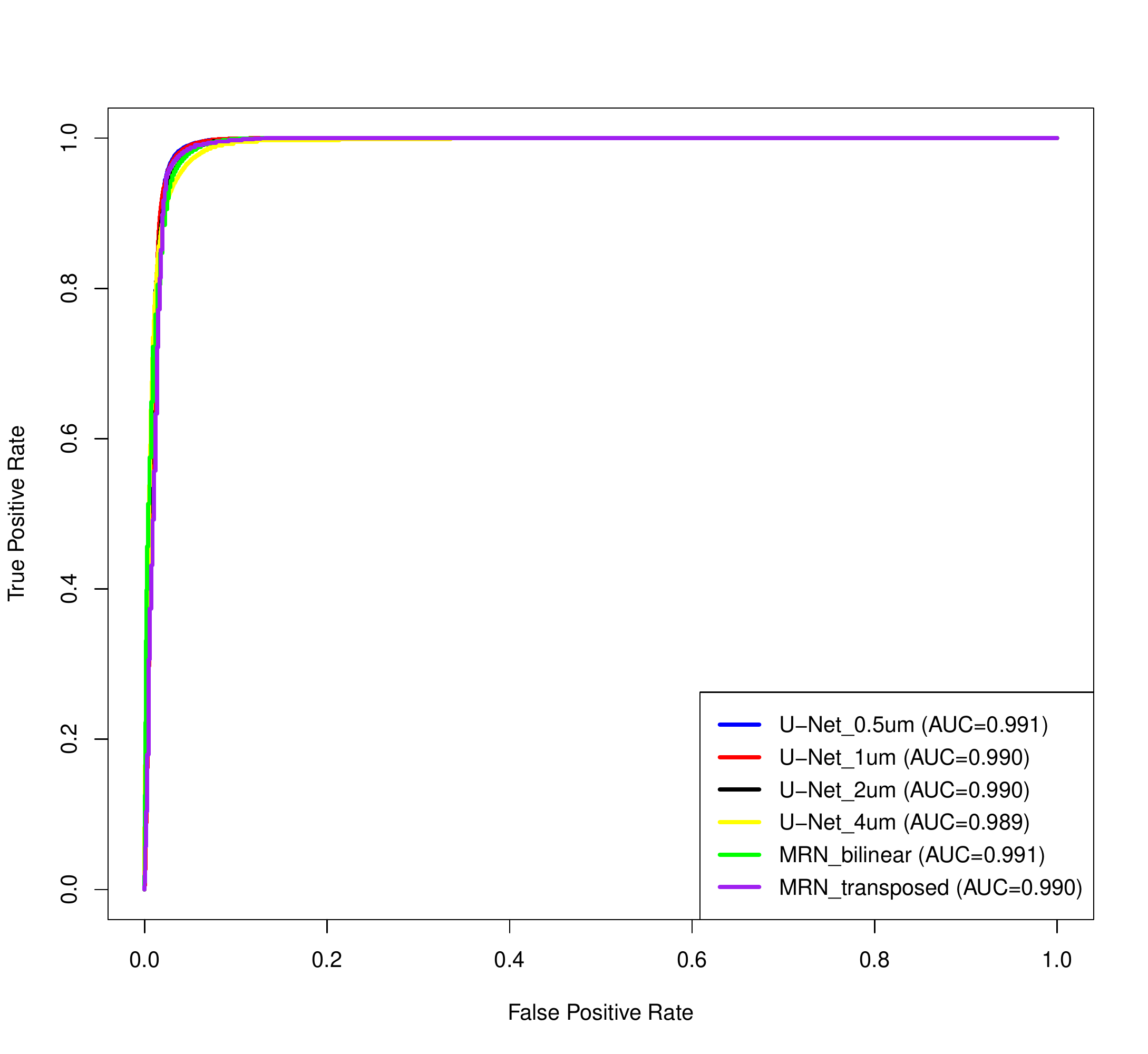}}
	\subfloat[Testing Set]{\includegraphics[width=0.5\textwidth]{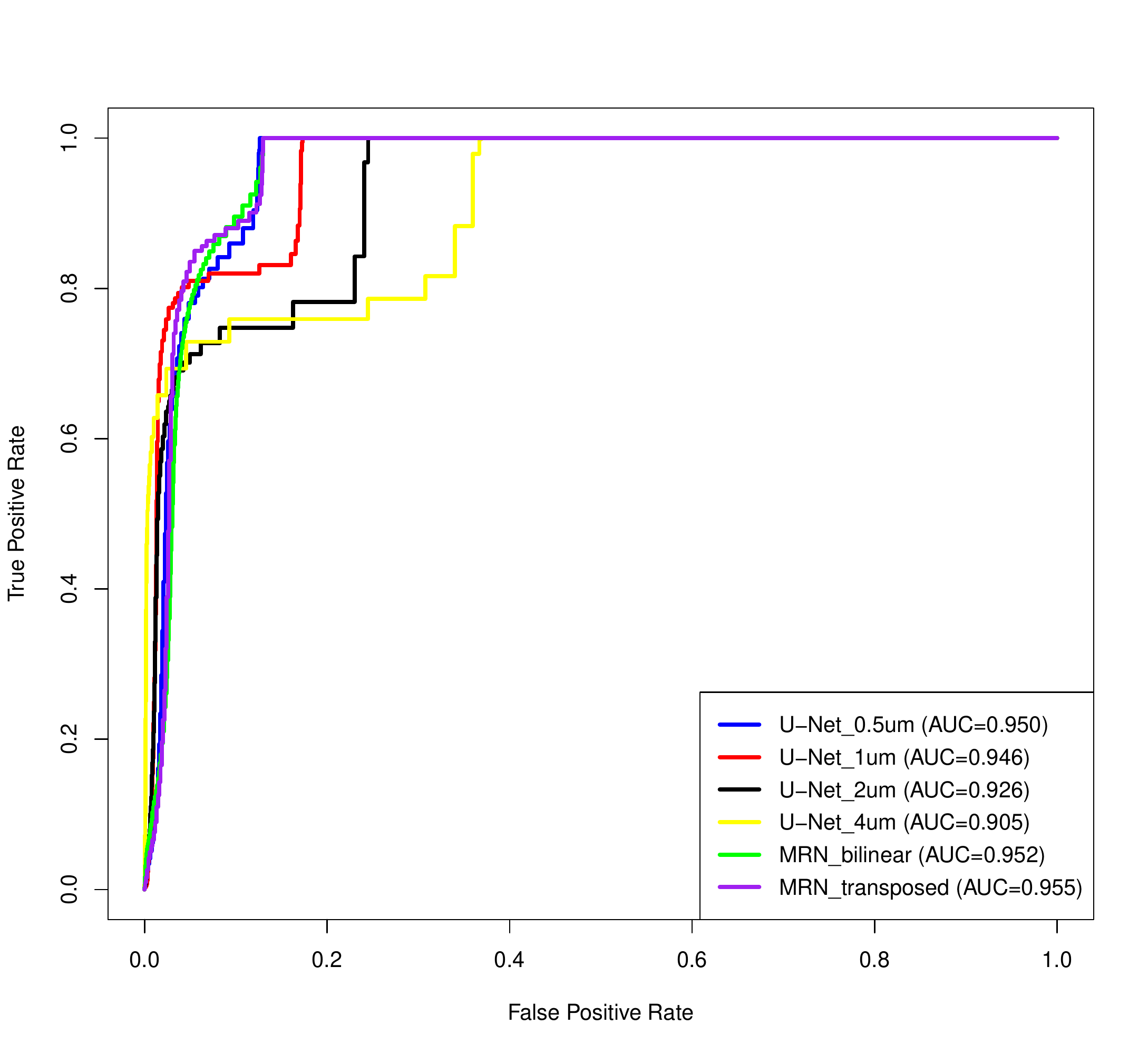}}
\caption{Comparisons of the standard U-Net and MRNs on CAMELYON16, when different thresholds are applied to the predictions.}
\label{fig:results_c16_test}
\end{figure}

\begin{figure}[t]
\centering
\includegraphics[width=\textwidth]{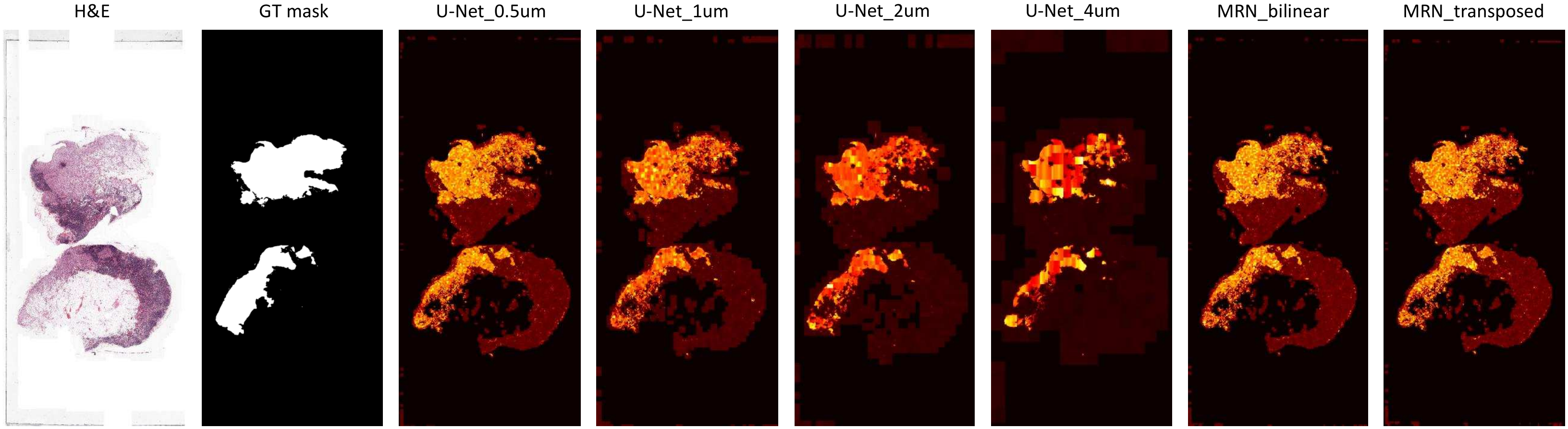}
\caption{A qualitative comparison of all methods on `test\_090' slide.}
\label{fig:test_021}
\end{figure}

\section{Results and Analysis}\label{sec:rets_analysis}

In this section, we compare the methods from both quantitative and qualitative perspectives. Quantitatively, we intend to evaluate their learning abilities on the training data, and more importantly the generalization capabilities on unseen data in the testing set. Therefore, we evaluated on both the validation set and the testing set, the ROC curves are displayed in Fig.~\ref{fig:results_c16_test}. On the validation set, the results are rather identical, and MRN-transposed is marginally better. This indicates that all methods are able to learn from the given data. Results of the testing set vary more significantly. First of all, the standard U-Net performance decreases as the mpp value increases, while the proposed networks both outperform the U-Net variants. The reason for the better performance of MRN-transposed can be that it has a higher capacity than MRN-bilinear, since transposed convolutions are parameterized and bilinear interpolations are not.  

To understand the results qualitatively, we plot the original H\&E slide, the annotation mask, and the predictions of trained models, as shown in Fig.~\ref{fig:test_021}. As the mpp value goes up, predictions of the U-Net variants become increasingly sparser and less confident (implied by darker colors). However, the predictions of both MRN-bilinear and MRN-transposed contain sufficient amount of details and are relatively more confident. This explains why they produce the best performance when evaluating at the pixel level.

\section{Conclusions and Future Work}\label{sec:con_future}
In this paper, we proposed two novel multiple resolution networks, to learn from and infer on WSIs at different resolutions. The proposed methods produce state-of-the-art results and outperform the standard U-Net on a benchmark dataset, for binary semantic segmentation. These results demonstrate their superior learning and generalization capabilities. In addition, the proposed methods are memory efficient, since constant input shapes of different resolutions make the increase in VRAM linear for training and prediction. Furthermore, we can now train one model for all resolutions of interest, instead of training one model for each.   

As for the future work, we would like to apply the proposed methods to other more challenging problems, e.g. multi-class semantic segmentation. Other network architectures can also be transformed to be multi-resolution capable, following the same principles proposed in this work. In addition, we will experiment with other building blocks of semantic segmentation networks, to develop methods with higher capacities.

\subsubsection*{Acknowledgements}

The authors would like to thank ContextVision AB, Sweden for supporting the research, and the organizers of CAMELYON challenges for making the datasets available to the community.

\bibliographystyle{splncs03}
\bibliography{paper}

\end{document}